\documentclass{article}
\usepackage{ijcai17}

\usepackage{times}

\usepackage{amsmath,amssymb,amsfonts,amsthm}
\usepackage{algorithm}
\usepackage{algorithmic}
\usepackage{graphicx}
\usepackage[small]{caption}
\usepackage{subcaption}
\usepackage{color}

\newcommand{\pr}{\mathrm{P}}
\newcommand{\eg}{\textit{e.g.}}
\newcommand{\ie}{\textit{i.e.}}
\newcommand{\wrt}{\textit{w.r.t.}}
\newcommand{\etal}{\textit{et al}}

\newcommand{\DP}{\mathrm{DP}}

\newcommand{\qset}{\mathcal{Q}}
\newcommand{\rset}{\mathcal{R}}
\newcommand{\sset}{\mathcal{S}}

\title{Scalable Estimation of Dirichlet Process Mixture Models on Distributed Data}
\author{Ruohui Wang\\ 
	Department of Information Engineering,\\
	The Chinese University of Hong Kong \\
	wr013@ie.cuhk.edu.hk \\
	\And
	Dahua Lin\\ 
	Department of Information Engineering,\\
	The Chinese University of Hong Kong \\
	dhlin@ie.cuhk.edu.hk
}

\begin{document}

\maketitle


\begin{abstract} 
We consider the estimation of Dirichlet Process Mixture Models (DPMMs) 
in distributed environments, where data are distributed across multiple computing nodes. 
A key advantage of Bayesian nonparametric models such as DPMMs is that they allow 
new components to be introduced on the fly as needed. 
This, however, posts an important challenge to distributed estimation -- 
how to handle new components efficiently and consistently. 
To tackle this problem, we propose a new estimation method, 
which allows new components to be created locally in individual computing nodes.
Components corresponding to the same cluster will be identified and merged 
via a probabilistic consolidation scheme.
In this way, we can maintain the consistency of estimation with very low communication cost.
Experiments on large real-world data sets show that the proposed method can achieve high scalability in distributed and asynchronous environments without compromising the mixing performance.
\end{abstract} 


\section{Introduction}
\label{intro}

\emph{Dirichlet Process Mixture Models (DPMMs)}~\cite{antoniak1974mixtures}
is an important family of mixture models,
which have received much attention from the statistical learning community
since its inception.
Compared to classical mixture models for which the number of components
has to be specified \textit{a priori}, DPMMs allow the model size to change as needed.
Hence, they are particularly suited to exploratory study, especially
in the contexts that involve massive amount of data.

Various methods have been developed for estimating DPMMs from data.
From earlier methods based on the \emph{Chinese Restaurant Process (CRP)}
formulation~\cite{maceachern1998estimating}
to recent ones that resort to merge-split steps~\cite{jain2004split}
or variational formulations~\cite{Blei05variationalinference},
the performance has been substantially improved.
Most of these methods adopt a serial procedure,
where updating steps have to be executed sequentially, one after another.
%
As we move steadily towards the era of big data, Bayesian nonparametrics,
like many other machine learning areas, is faced with a significant challenge,
namely, to handle \emph{massive data sets} that may go beyond the capacity
of a single computing node. Tackling such a challenge requires new techniques
that can process different parts of the data concurrently.
However, most existing methods for DPMM estimation adopt an iterative procedure,
and therefore they are not able to scale in a distributed environment.

In recent years, parallel methods~\cite{williamson2013parallel,chang2013parallel}
have been developed, which attempt to speed up the estimation of DPMMs through
parallel processing, by exploiting the conditional independence of the model.
Note that these parallel methods are based on the \emph{shared memory architecture},
where the entire dataset together with the intermediate results are held
in a unified memory space, and all \emph{working threads} can access them
without costly communication.
%
However, in large-scale applications, the amount of data can go far beyond
the capacity of a single computer.
Handling such data requires a \emph{distributed architecture},
where multiple computers, each called a \emph{computing node},
are connected via communication channels with limited bandwidth,
\eg~\emph{Ethernet}. Computing nodes do not share memory --
information exchange has to be done via communication.
The parallel methods mentioned above, when applied to such settings,
would incur considerable communication costs.
For example, changing associations between samples and processors
can result in frequent data transfer.

In this work, we aim to develop a new method for DPMM estimation that
can scale well on a \emph{distributed computing architecture}.
Distributed estimation is not a new story -- a variety of methods
for estimating \emph{parametric models} from distributed data~\cite{newman2009distributed}
have been developed in recent years.
However, \emph{nonparametric models} like DPMMs, present additional challenges
due to the possibility of new components being introduced on the fly.
Therefore, how to handle new components efficiently and consistently
becomes an important issue. On one hand, to attain high concurrency,
one has to allow local workers to discover new components independently;
on the other hand, components arising from different workers may actually
correspond to the same cluster, which need to be identified and merged
in order to form a coherent estimation.
The trade-off between mixing performance and communication cost
is also an important issue.

In tackling this problem, we develop a distributed sampling algorithm,
which allow new components to be introduced by local workers,
while maintaining the consistency among them through two
consolidation schemes, namely \emph{progressive consolidation} and
\emph{pooled consolidation}.
We tested the proposed methods on both synthetic and large real-world
datasets. Experimental results show that they can achieve reasonably high 
scalability while maintaining the convergence speed.
It is also worth noting that the proposed method can work under
asynchronous settings without performance degradation.


\section{Related Work}

With the rapid growth of data,
parallel and distributed methods have received increasing attention.
Earlier efforts along this line focused on the estimation of parametric models.
Newman \etal~\cite{newman2007distributed,newman2009distributed} presented a method
for estimating LDA models~\cite{blei2003latent} on distributed data,
where concurrent sampling on local subsets of data are followed by
a global update of the topic counts.
Smyth~\etal~\cite{smyth2009asynchronous} further extend this algorithm to asynchronous settings.
All these methods assume a fixed parameter space, and therefore they can not be
directly used for estimating Bayesian nonparametric models,
of which the size of the parameter space can vary on the fly.

For DPMMs, a variety of estimation methods based on different
theoretical foundations have been developed, such as
Chinese Restaurant Process~\cite{maceachern1998estimating},
stick-breaking reconstruction~\cite{sethuraman1994constructive},
Poisson processes~\cite{lin2010construction},
and slice sampling~\cite{walker2007sampling}.
The serial nature of these methods make them difficult to be parallelized.

Driven by the trend of concurrent computing, recent years witnessed new efforts
devoted to parallel estimation of BNP models.
Chang and Fisher~\cite{chang2013parallel} proposed an MCMC algorithm that
accomplishes both intra-cluster and inter-cluster parallelism by
augmenting the sample space with super-cluster groups.
In the same year, Williamson~\etal~\cite{williamson2013parallel} proposed
another parallel sampler for DPMMs, which exploits the conditional independence
among components through auxiliary weights.
Assuming that all processors share memory, both methods update associations
between samples and processors in each iteration. Hence, they are not suitable for
distributed estimation where information can only be exchanged across limited
communication channels.
Also, it has been found~\cite{gal2014pitfalls} that some parallel methods
such as~\cite{williamson2013parallel} have the issue of unbalanced workload among processors.

Recently, attempts have been made to further extend BNP estimation to distributed environments.
Ge \etal~\cite{ge2015distributed} developed a distributed estimation method for DPMMs
based on the slice sampler presented in~\cite{walker2007sampling}.
This method adopts a map-reduce paradigm, where the \emph{map} step is to
sample component labels, while the \emph{reduce} step is to accumulate component weights,
update parameters, or create new components.
This method has a limitation: new components are sampled from the prior
without reference to the observed data, often yielding poor fits.
In our experiments, we observed that it often converges very slowly.
Also note that unlike ours, this method cannot operate in asynchronous modes.
Campbell \etal~\cite{campbell2015streaming} proposed a variational inference method
that targets streaming and distributed contexts.
This method explicitly merges components from each mini-batch to
a central pool by solving a combinatorial problem.
An important issue of this method is that it lacks a splitting step
to revert undesirable merges.
Newman \etal's paper~\cite{newman2009distributed} also described a nonparametric
extension to their method. This extended method merges topics from different
workers simply by topic-ids or greedy matching, thus often yielding incorrect mergers.



\section{Dirichlet Process Mixture Models}
\label{sec:background}

%
A \emph{Dirichlet Process (DP)}, denoted by $\DP(\alpha \mu)$,
is a stochastic process characterized by a \emph{concentration parameter} $\alpha$
and a \emph{base distribution} $\mu$.
A DP sample is almost surely discrete, and can be expressed as
$D = \sum_{k=1}^\infty \pi_k \delta_{\phi_k}$.
Here, each atom $\phi_k$ is associated with a weight $\pi_k$, which satisfies $\sum_k \pi_k = 1$.
\emph{DPMM} is a mixture model formulated on top of a DP, where the atoms $\{\phi_k\}$
serve as the component parameters:
\begin{equation} \label{eq:dpmm}
	D \sim \DP(\alpha \mu), \
	\theta_i \sim D, \
	x_i \sim F(\theta_i), \   i = 1, \ldots, n.
\end{equation}
Here, $F(\theta_i)$ indicates a generative component with parameter $\theta_i$,
which must be one of the atoms in $\{\phi_k\}$.
Whereas $D$ has infinitely many atoms,
only a finite subset of them are associated with the observed data.
%
A key advantage of DPMM as opposed to classical mixture models is that
the number of components $K$ need not be specified in advance.
Instead, it allows new components to be introduced on the fly.

%
Generally, a DPMM can be estimated based on the \emph{Chinese Restaurant Process},
an alternative characterization where $D$ is marginalized out.
Particularly, an indicator $z_i$ is introduced to attach the sample $x_i$ to
a certain component $\phi_k$ (with $k = z_i$).
Then, the estimation can be accomplished by alternating between
the sampling of $z_i$ and $\phi_k$.
%
%
In this paper, we focus on the case where
the prior $\mu$ is conjugate to the likelihood $f$.
Thus $f$ and $\mu$ can generally be written as:
\begin{equation} \label{eq:conjugate}
\begin{split}
	f(x | \phi) &=
	h(x) \, \exp \left( \eta(\phi)^T \psi(x) - c \cdot a(\phi) \right),
	\\
	\mu(\phi | \beta_0, \kappa_0) &=
	\exp \left(\beta_0^T \eta(\phi) - \kappa_0 \cdot a(\phi) - b(\beta_0, \kappa_0) \right).
\end{split}
\end{equation}

With this assumption, the posterior distribution of $\phi_k$, denoted by $\tilde{p}_k$,
is in the same \emph{exponential family} as $\mu$,
whose canonical parameters are given by
$\beta|_{S^{(k)}} = \beta_0 + \psi(S^{(k)})$ and
$\kappa|_{S^{(k)}} = \kappa_0 + c \cdot |S^{(k)}|$.
Here,
$S^{(k)}$ denotes the set of samples assigned to the $k$-th cluster,
and $\psi(S^{(k)})=\sum_{x\in S^{(k)}}\psi(x)$.
With conjugacy, the atoms $\phi_k$ can be easily marginalized out,
resulting in a more efficient scheme, called \emph{Collapsed Gibbs Sampling (CGS)},
which iteratively applies a \emph{collapsed step}:
\begin{equation} \label{eq:sassign1}
	P(z_i = k | z_{/i}, X) \propto
	\begin{cases}
		n_{/i}^{(k)} \bar{f}(x_i | \tilde{p}_{k/i}) & (1 \le k \le K)
		\\
		\alpha \bar{f}(x_i | \mu) & (k = K+1)
	\end{cases}.
\end{equation}
Here, $n_{/i}^{(k)}$ is the number of samples assigned to the $k$-th cluster (except $x_i$),
and $\bar{f}(x_i | p)$ is a marginal density \wrt~$p$,
given by $\bar{f}(x | p) = \int f(x | \theta) p(d\theta)$,
which has an analytic form $ 
h(x) \exp \left(
b(\beta_p + \psi(x), \kappa_p + c) -
b(\beta_p, \kappa_p)
\right)$ under the conjugacy assumption~\cite{blei2016exponential}.


\section{Distributed Estimation}

Towards the goal of developing a scalable algorithm for estimating DPMMs
in a \emph{distributed} environment, we are faced with two challenges:
(1) High scalability requires computing nodes to work independently
without frequent communication. This requirement, however, is confronted by the
extensive dependencies among samples due to the marginalization of $D$.
(2) To increase concurrency, it is desirable to allow individual workers
to create new components locally. This, however, would lead to the issue of
\emph{component identification}, \ie~new components from different workers
can correspond to the same cluster.
Our method tackles these challenges by allowing individual workers to
update parameters or create new components \emph{independently},
while enforcing consistency among them by a delayed
\emph{consolidation} stage.


\subsection{Synchronizing Existing Components}
\label{sub:delaysync}

Suppose we have $M$ local workers and a master node.
The sample set $X$ is partitioned into disjoint sets $X_1, \ldots, X_M$,
each in a worker. Both the sample count $n^{(k)}$ and the sufficient statistics $\psi(S^{(k)})$ can be computed by summing up
their local counterparts, as $n^{(k)} = \sum_{l=1}^M n_l^{(k)}$ and
$\psi(S^{(k)}) = \sum_{l=1}^M \psi(S_l^{(k)})$,
where $S_l^{(k)}$ is the set of samples in the $l$-th worker that are assigned
to the $k$-th cluster, and $n_l^{(k)} = |S_l^{(k)}|$.

When a sample is reassigned, \ie~$z_i$ changes,
relevant statistics need to be updated, which would incur frequent communication.
We address this via \emph{delayed synchronization}.
Specifically, the master node maintains a \emph{global version} of the estimated parameters,
denoted by $\{(\beta_g^{(k)}, \kappa_g^{(k)})\}$,
while each worker maintains a \emph{local version}.
At each cycle, the worker fetches the latest \emph{global version}
from the master node, and then launches local updating as presented in Eq.\eqref{eq:sassign1}.
At the end of a cycle, the worker pushes the \emph{deltas},
\ie~the differences between the updated parameters and the fetched versions,
to the master.
Take a closer look. At the beginning of a cycle,
each worker (say the $l$-th) obtains a global version from the master,
where the parameters are given by
\begin{equation}
	\beta_g^{(k)} = \beta_0 + \sum_{l=1}^M \phi(S_l^{(k)}), \
	\kappa_g^{(k)} = \kappa_0 + c \cdot \sum_{l=1}^M |S_l^{(k)}|.
\end{equation}
After sample re-assignment, the $k$-th cluster changes from
$S_l^{(k)}$ to ${S'}_l^{(k)}$, thus the local parameters will be updated to
\begin{align}
	{\beta'}_l^{(k)} &= \beta_0 + \phi({S'}_l^{(k)}) + \sum_{j \ne l} \phi(S_j^{(k)}), \notag \\
	{\kappa'}_l^{(k)} &= \kappa_0 + c \cdot |{S'}_l^{(k)}| + c \cdot \sum_{j \ne l} |S_j^{(k)}|.
\end{align}
Then the \emph{deltas} from worker $l$ would be
\begin{align}
	\Delta \beta_l^{(k)}
	&= {\beta'}_l^{(k)} - \beta_g^{(k)} = \phi({S'}_l^{(k)}) - \phi(S_l^{(k)}), \notag \\
	\Delta \kappa_l^{(k)}
	&= {\kappa'}_l^{(k)} - \kappa_g^{(k)} = c \cdot (|{S'}_l^{(k)}| - |{S}_l^{(k)}|).
\end{align}
When receiving such deltas from all local workers, the master would add them to the global version.
Provided that \emph{no new components are created in this cycle},
the updated global version would exactly match the new sample assignments.

\emph{Delayed synchronization} is an \emph{approximation},
which trades \emph{mathematical rigorousness} for \emph{high scalability}.
As shown in our experiments, it has little impact on the convergence performance,
but substantial influence on scalability.


\subsection{Consolidating New Components}
\label{sub:consol_new}

Local updates can create new components --
new components from different workers may correspond to the same cluster.
It is important to identify such components and merge them, as treating
them as different would lead to misleading estimates.
This is an important challenge in our work.

The identity between components can be determined via hypothesis testing.
Given a set of samples $X$ and a collection of clusters $\{S_1, \ldots, S_K\}$.
The first hypothesis, denoted by $H_0$, is that these clusters are from different components;
while the alternative one, denoted by $H_1$, is that $S_1$ and $S_2$ are from the same one.
%
With the DPMM formulation in Eq.\eqref{eq:dpmm} and the conjugacy assumption in Eq.\eqref{eq:conjugate}, we have
\begin{equation} \label{eq:msratio}
	\begin{split}
		&\frac{\pr(H_1 | X)}{\pr(H_0 | X)}
		=
		\frac{\pr(H_1)}{\pr(H_0)} \cdot
		\frac{\pr(X | H_1)}{\pr(X | H_0)}
		\\
		&=\frac{1}{\alpha} 
		\frac{\Gamma(|S_{12}|)}{\Gamma(|S_1|) \Gamma(|S_2|)} \cdot
		\frac
		{\exp\left(b(\beta_{12}, \kappa_{12}) + b(\beta_0, \kappa_0)\right)}
		{\exp\left(b(\beta_1, \kappa_1) + b(\beta_2, \kappa_2) \right)}.
	\end{split}
\end{equation}
Here,
$S_{12} \triangleq S_1 \cup S_2$ with $|S_{12}| = |S_1| + |S_2|$,
$\beta_k \triangleq \beta_0 + \psi(S_k)$, and
	$\kappa_k \triangleq \kappa_0 + c \cdot |S_k|$.
%
%
In what follows, we will refer to the ratio given by Eq.\eqref{eq:msratio} as the \emph{merge-split ratio} of $(S_1, S_2)$, and denote it by $\rho(S_1, S_2)$.
Note that computing $\rho(S_1, S_2)$ requires only the sufficient statistics of $S_1$ and $S_2$,
and therefore it can be done by the master node without the need to access the data.
Based on this, we derive two schemes to handle new components:
\emph{Progressive consolidation} and \emph{Pooled consolidation}.
Note in following sections , we mix the use of set symbol $S$ and its corresponding statistics $(\beta,\kappa)$. As in consolidation operations, sample sets are treated as whole and calculation involves statistics only.

\subsubsection{Progressive Consolidation}
\label{subsub:prog}

As mentioned, the master maintains the global versions of the canonical parameters
$\{(\beta_g^{(k)}, \kappa_g^{(k)})\}$, and will receive the
\emph{deltas} $\{(\Delta \beta_l^{(k)}, \Delta \kappa_l^{(k)})\}$ from local workers.
The \emph{Progressive Consolidation} scheme incorporate the deltas one by one.
Particularly, the \emph{delta} from a worker may comprise two parts:
\emph{updates to existing components} and \emph{new components}.
The former can be directly added to the global version as discussed in
Sec~\ref{sub:delaysync}; while the latter can be incorporated via \emph{progressive merge}.
To be more specific, given a new component $(\beta', \kappa')$,
the master has $K+1$ choices, merging it with either of the $K$ existing components
or adding it as the $(K+1)$-th one.
The posterior probabilities of these choices can be computed based on Eq.\eqref{eq:msratio}:
\begin{equation} \label{eq:prog_choice}
	\pr(u = k | X) \propto
	\begin{cases}
		\rho(S^{(k)}_g, S') & (1 \le k \le K), \\
		1 & (k = K + 1).
	\end{cases}
\end{equation}
Here, $u$ indicates the choice -- when $u = k \le K$, the new component is merged to the $k$-th one,
and when $u = K + 1$, the new component is added as a new one.
Key steps of \emph{progressive consolidation} are summarized in Algorithm~\ref{alg:prog}.

\begin{algorithm}[t]
	\caption{Progressive Consolidation}
	\label{alg:prog}
	\begin{algorithmic}
		\STATE {\bfseries Given:} \\
		$\quad$ Global collection: $\qset=\{S_g^{(1)},\ldots,S_g^{K}\}$, \\
		$\quad$ Deltas $\{\Delta_l\}_{l=1}^M$
		where $\Delta_l = \{(\Delta \beta_l^{(k)}, \Delta \kappa_l^{(k)})\}$
		\FOR{$l=1$ {\bfseries to} $M$}
		\FOR{$k=1$ {\bfseries to} $K$}
		\STATE $\beta_g^{(k)} \leftarrow \beta_g^{(k)} + \Delta \beta_l^{(k)}, \
		\kappa_g^{(k)} \leftarrow \kappa_g^{(k)} + \Delta \kappa_l^{(k)}$
		\ENDFOR
		\FOR{$k'=K+1$ {\bfseries to} $|\Delta_l|$}
		\STATE Compute $\rho(S_g^{(k)}, S_l^{(k')})$ for $k = 1, \ldots, |\qset|$
		\STATE Draw $u \in \{1, \ldots, |\qset|+1\}$ as Eq.\eqref{eq:prog_choice}
		\IF {$u \le |\qset|$}
		\STATE $\beta_g^{(u)} \leftarrow \beta_g^{(u)} + \Delta \beta_l^{(k')}$
		\STATE $\kappa_g^{(u)} \leftarrow \kappa_g^{(u)} + \Delta \kappa_l^{(k')}$
		\ELSE
		\STATE $\qset \leftarrow \qset \cup \{(\Delta \beta_l^{(k')}, \Delta \kappa_l^{(k')})\}$
		\ENDIF
		\ENDFOR
		\ENDFOR
	\end{algorithmic}
\end{algorithm}

\subsubsection{Pooled Consolidation}
\label{subsub:pool}

\emph{Progressive consolidation} has a limitation:
it takes very long to correct a wrong merger --
wait until new components to take the place of the wrongly merged one.
To address this issue, we propose an MCMC algorithm called \emph{Pooled Consolidation}.
This algorithm pools all local updates and consolidates them altogether
via \emph{merge} and \emph{split} steps.
Specifically, this algorithm has multiple iterations,
each proposing a \emph{merge} or a \emph{split}, with equal chance.

\vspace{-10pt}
\paragraph{Merge proposal.}
Generally, components with high merge-split ratios are good candidates for a merge.
To propose a merge step, we choose a pair of distinct components $A$ and $B$ from global collection $\qset$, with a probability proportional to the \emph{merge-split ratio} $\rho(A, B)$.
To facilitate splitting, we will keep track of the set of \emph{sub-components} for each component $A$, denoted by $\sset_A$.
Components that are created by local updates are \emph{atomic} and cannot be split. For an atomic component $A$, $\sset_A = \{A\}$. When two components $A$ and $B$ are merged into a \emph{non-atomic} one $C$, we have $\sset_C = \sset_A \cup \sset_B$.

\vspace{-10pt}
\paragraph{Split proposal.}
For a non-atomic component $C$, there are $2^{|\sset_C| - 1} - 1$ ways to split it into two.
Hence, finding a reasonable split is nontrivial.
Our idea to tackle this problem is to \emph{unpack and re-consolidate} the sub-components
in $\sset_C$ using a restricted version of the progressive consolidation.
Particularly, we begin with an empty collection $\rset$, and progressively merge
sub-components in $\sset_C$ to $\rset$. When $|\rset|$ reaches $2$, all remaining sub-components
can only be merged into either element of $\rset$, \ie~they cannot be added as new components.
This will yield either a single component, which is just $C$, or two components.
Let $\sset_C = \{A_1, \ldots, A_m\}$.
The probability that this would end-up with a single-component is:
\begin{equation}
	\beta_C \triangleq \prod_{j=1}^{m-1} \frac{\rho(A_{1:j}, A_{j+1})}{\rho(A_{1:j}, A_{j+1}) + 1}.
\end{equation}
Here, $A_{1:j}$ denotes a component that combines $A_1, \ldots, A_j$.
%
Generally, small values of $\beta_C$ tend to indicate a good candidate for splitting.
To propose a \emph{split}, we choose a non-atomic component $C$, with a probability proportional to $1 / \beta_C$, and generate a split $(A, B)$ via \emph{Restricted Consolidation} as described above (Algorithm~\ref{alg:rprog} shows the detailed steps).
Note that the probability of the resultant split, denoted by $\gamma_C(A, B)$, can be computed by taking the products of the probabilities of all choices made along the way.

\begin{algorithm}[t]
	\caption{Restricted Consolidation}
	\label{alg:rprog}
	\begin{algorithmic}
		\STATE {\bfseries Input:} A set of atomic components:
		$\sset_C =\{A_1,\ldots,A_m\}$. \\
		\STATE Initialize $R_1 = A_1$, $\rset = \{R_1\}$, $\gamma_C = 1$
		\FOR{$k=2$ {\bfseries to} $m$}
		\STATE Compute $w_i = \rho(R_i, A_k)$ for each $R_i \in \rset$
		\STATE Set $w_2 = 1$ if $|\rset| = 1$
		\STATE $p_i \leftarrow w_i / (w_1 + w_2)$ for $i = 1, 2$
		\STATE Draw $u \in \{1, 2\}$ with $\pr(c = i) = p_i$
		\STATE \textit{\# Progressively compute the split probability:}
		\STATE $\gamma_C \leftarrow \gamma_C \cdot p_u$
		\IF {$u \le |\rset|$}
		\STATE \textit{\# Merge to component $R_u$:}
		\STATE $\beta_r^{(u)} \leftarrow \beta_r^{(u)} + \psi(A_k), \
		\kappa_r^{(u)} \leftarrow \kappa_r^{(u)} + c \cdot |A_k|$
		\ELSE
		\STATE \textit{\# Add as the second component $R_2$:}
		\STATE $\rset \leftarrow \rset \cup \{(\beta_0 + \psi(A_k), \kappa_0 + c \cdot |A_k|)\}$
		\ENDIF
		\ENDFOR
		\STATE {\bfseries Output:} The resultant split $\rset$ and the probability $\gamma_C$.
	\end{algorithmic}
\end{algorithm}

\vspace{-10pt}
\paragraph{Acceptance probability.}
From the standpoint of MCMC, merging $A$ and $B$ into $C$ is
a move from $\qset$ to $\qset' = (\qset - \{A, B\}) \cup \{C\}$,
while the step of splitting $C$ into $A$ and $B$ reverses this.
Based on the proposal procedure described above, we derive the transition probabilities:
\begin{equation}
	\pr(\qset \rightarrow \qset')
	= \frac{\rho(A, B)}{\sum_{A \ne B} \rho(A, B)},
\end{equation}
\begin{equation}
	\pr(\qset' \rightarrow \qset)
	= \frac{1 / \beta_C}{\sum_{C' \in \qset'} (1 / \beta_{C'})} \gamma_C(A, B).
\end{equation}
Note that $\pr(\qset | X) / \pr(\qset' | X) = \rho(A, B)$.
Consequently, the acceptance probabilities are given by
\begin{equation}
	a((A, B) \rightarrow C)
	= \min\left(1, \rho(A, B)
	\frac{\pr(\qset' \rightarrow \qset)}
	{\pr(\qset \rightarrow \qset')}
	\right),
\end{equation}
\begin{equation}
	a(C \rightarrow (A, B))
	= \min\left(1, \frac{1}{\rho(A, B)}
	\frac{\pr(\qset \rightarrow \qset')}
	{\pr(\qset' \rightarrow \qset)}
	\right).
\end{equation}
%


\subsection{Asynchronous Algorithm}
\label{sub:async}

Both \emph{progressive consolidation} and \emph{pooled consolidation} can be readily extended
to an \emph{asynchronous} setting,
where each worker has its own \emph{local schedule}.
At the end of a local cycle, the worker pushes \emph{deltas} to the master, pulls
the latest global version, and launch the next local cycle \emph{immediately} thereafter.
While the workers are doing their local updates,
the master can perform \emph{merging} and \emph{splitting} (Algorithm~\ref{alg:rprog}) concurrently
to refine the global pool -- the refined version will be available to the local workers in the next cycle.
%


\section{Experiments}
\label{experiments}

\paragraph{Datasets and Models.}
We evaluated the proposed methods on three datasets, a synthetic one and
two large real-world datasets: \emph{ImageNet} and \emph{New York Times Corpus}.

\begin{figure}[t]
	\centering
	\includegraphics[width=\columnwidth]{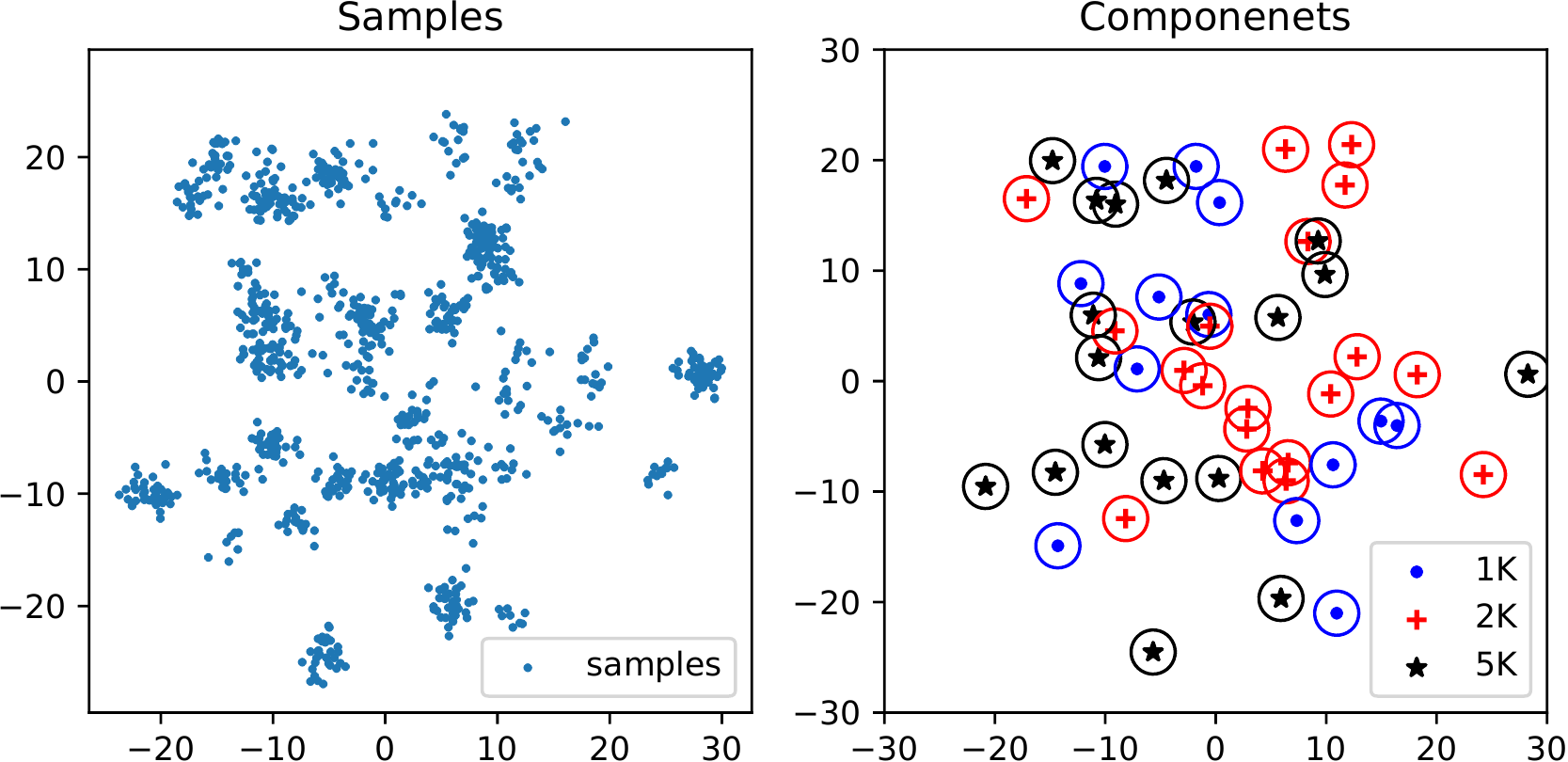}
	\caption{Visualization of the synthetic data set}
	\label{fig:syn_visual}
\end{figure}

The synthetic dataset is for studying the behavior of the proposed method.
It comprises $141K$ two-dimensional points from $50$ Gaussian components with unit variance.
The sizes of clusters range from $1000$ to $5000$ to emulate the unbalanced settings
that often occur in practice.
Figure~\ref{fig:syn_visual} shows the data set.
We can observe overlaps among clusters, which makes the estimation nontrivial.

The \emph{ImageNet} dataset is constructed from the training set of ILSVRC~\cite{ILSVRC15},
which comprises $1.28M$ images in $1000$ categories.
We extract a $2048$-dimensional feature for each image with
Inception-ResNet~\cite{DBLP:journals/corr/SzegedyIV16} and reduce the dimension to $48$ by PCA.
Note that our purpose is to investigate mixture modeling
instead of striving for top classification performance.
Hence, it is reasonable to reduce the feature dimension to a moderate level,
as samples are too sparse to form clusters in a very high dimensional space.
We formulate a Gaussian mixture to describe the feature samples, where
the covariance of each Gaussian components is fixed to $\sigma^2 I$ with
$\sigma = 8$. We use $\mathcal{N}(0, \sigma_0^2 I)$ as the prior distribution
over the mean parameters of these components, where $\sigma_0 = 8$.

For the \emph{New York Time (NYT) Corpus}~\cite{sandhaus2008new},
we construct a vocabulary with $9866$ distinct words, and derive a
bag-of-word representation for each article. Removing those with
less than $20$ words, we obtain a data set with about $1.7M$ articles.
We use a mixture of multinomial distribution to describe the NYT corpus.
The prior here is a symmetric Dirichlet distribution with hyperparameter $\gamma = 1$.

\vspace{-10 pt}
\paragraph{Experiment Settings.}
We compared eight methods.
Four baselines:
\textbf{CGS} - Collapsed Gibbs sampling~\cite{neal2000markov},
\textbf{SliceMR} - Map-reduce slice sampler~\cite{ge2015distributed},
\textbf{AV} - Auxiliary variable parallel Gibbs sampler~\cite{williamson2013parallel}
\footnote{We improved the performance of \textit{AV}
by adding our consolidation scheme to its local inference steps,
which can effectively merge similar clusters assigned to the same processor during global steps.}, and
\textbf{SubC} - Parallel sampler via sub-cluster splits~\cite{chang2013parallel}.
Three different configurations of the proposed method:
\textbf{Prog} - Synchronous Progressive consolidation (Sec~\ref{sub:consol_new}),
\textbf{Pooled} - Synchronous Pooled consolidation (Sec~\ref{sub:consol_new}), and
\textbf{Async} - Asynchronous consolidation (Sec~\ref{sub:async}).
And 
\textbf{Hung} - we replace our consolidation step with Hungarian algorithm, which was adopted for  component identification in~\cite{campbell2015streaming}.

These algorithms were examined from different aspects,
including \emph{convergence}, \emph{clustering accuracy}, \emph{communication cost},
and \emph{scalability}.
Each algorithm was launched for $10$ times (with different random seeds) on all data sets.
We report the average of the performance metrics.
We conducted the experiments using up to $30$ \emph{workers} on multiple physical servers.
They can communicate with each other via Gigabit Ethernet or TCP loop-back interfaces.

\begin{figure}[t]
	\includegraphics[width=\columnwidth]{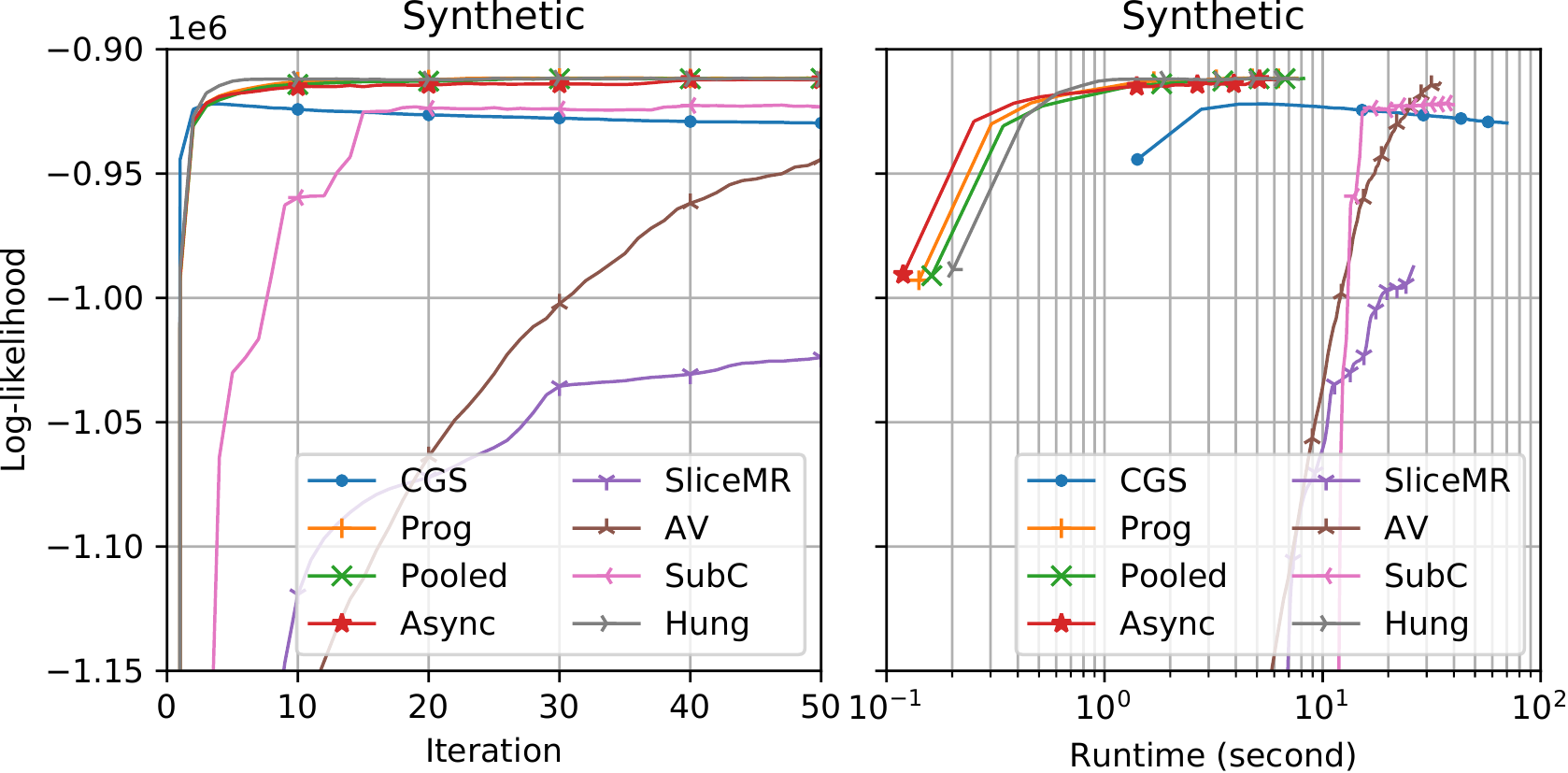}
	\includegraphics[width=\columnwidth]{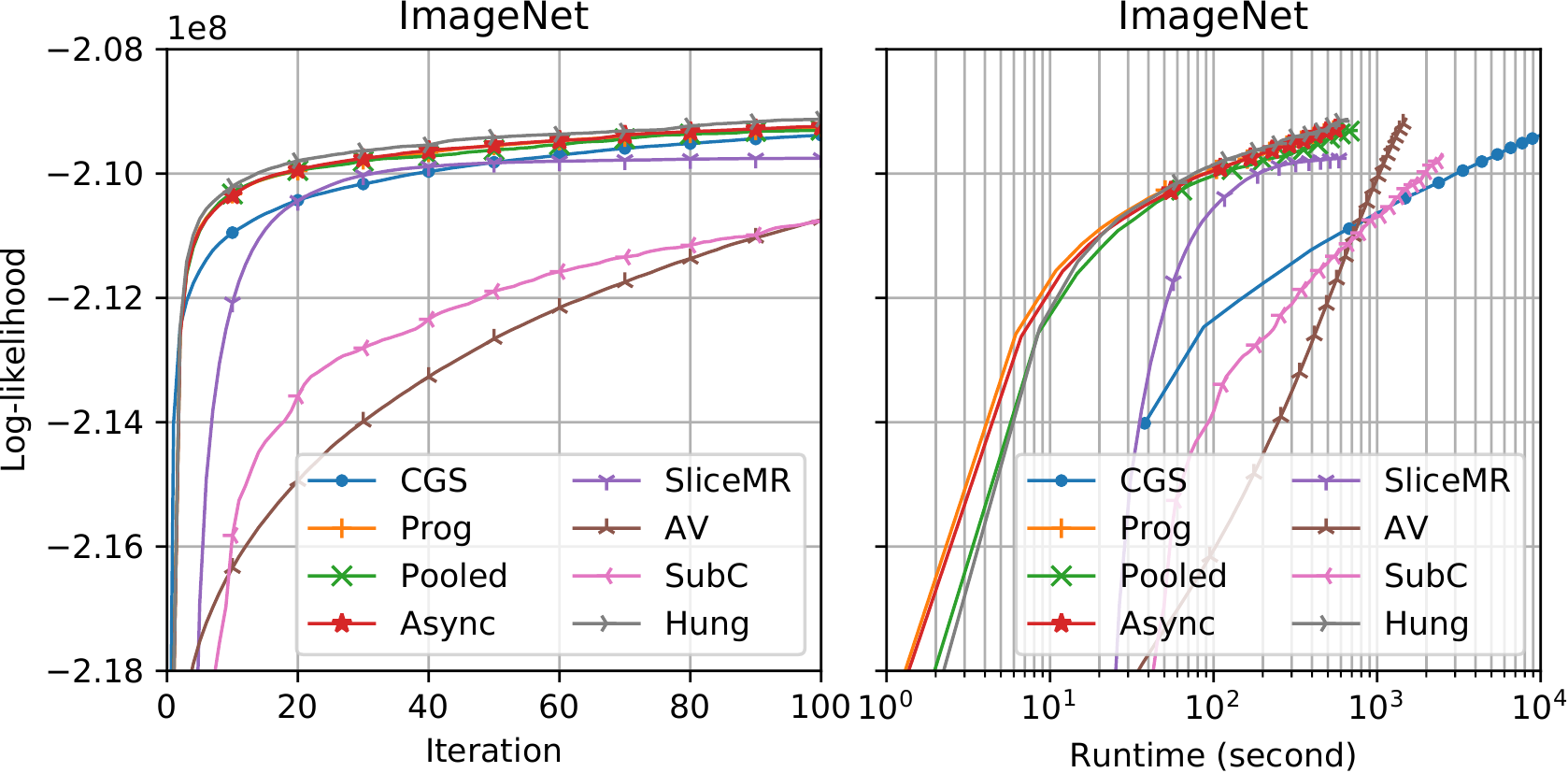}
	\includegraphics[width=\columnwidth]{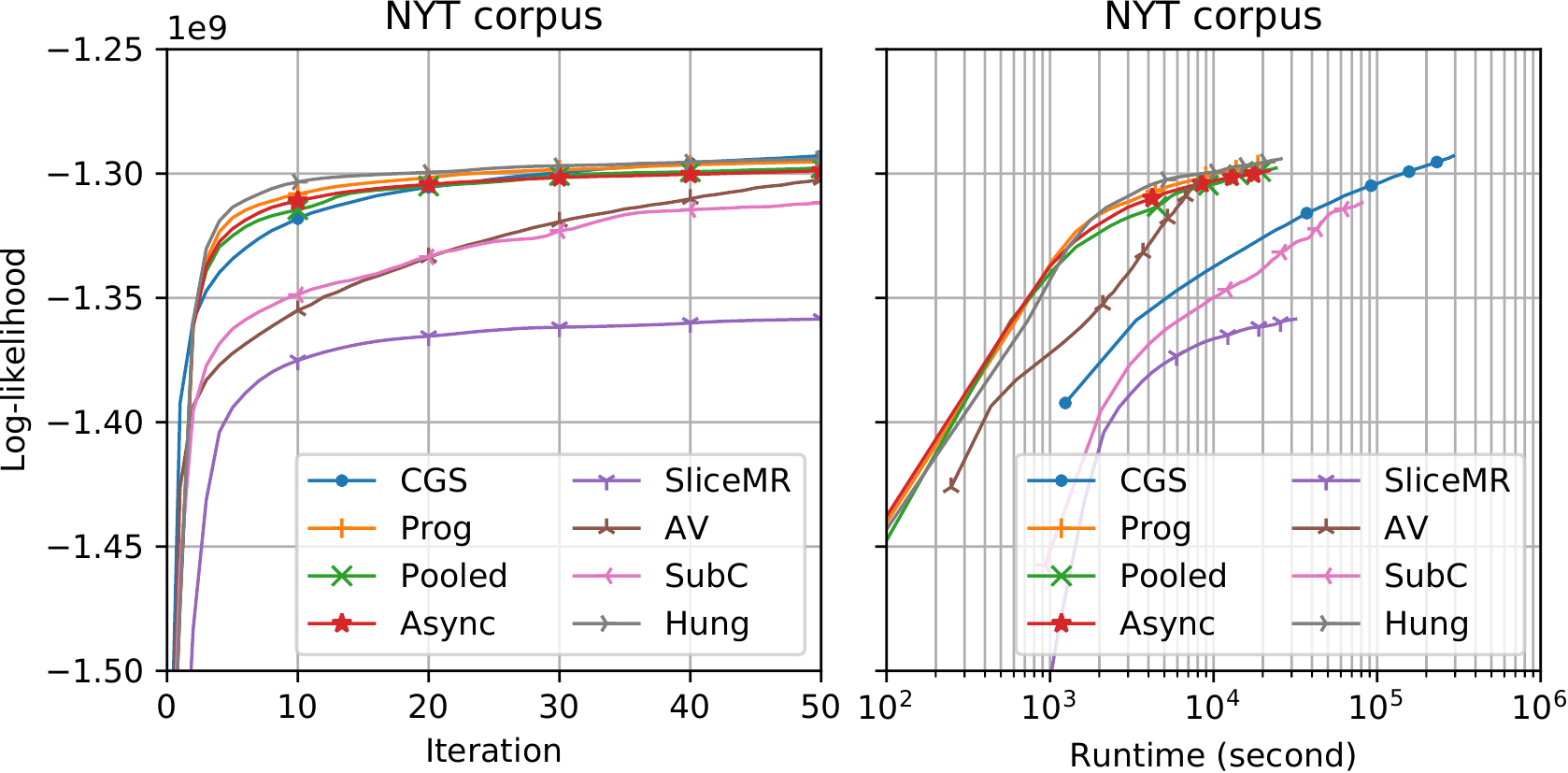}
	\caption{Log-likelihood \wrt iteration and runtime}
	\label{fig:likelihood}
\end{figure}

\vspace{-10 pt}
\paragraph{Convergence.}
We first compare the convergence of the log-likelihood.
Results on all three datasets are shown in Figure~\ref{fig:likelihood},
where the likelihoods are plotted as functions of
the number of iterations or the wall-clock time.

We observe that our algorithms can converge to the same level as \textit{CGS}
within comparable numbers of iterations on all three datasets,
but are about $10$ to $20$ times faster owing to concurrency.
On the synthetic dataset, we noticed that \textit{CGS} yields small noisy clusters,
which slightly decreased the likelihood.
While in our algorithms, such clusters will be merged via consolidation, thus resulting
in slightly higher likelihood.
Overall, our methods achieve high scalability without compromising the convergence performance.

Other baseline methods designed for parallel estimation,
namely \textit{SliceMR}, \textit{AV} and \textit{SubC},
usually take more iterations and thus longer runtime to converge.
Particularly,
in \textit{SliceMR}, new components are created from the prior distribution
without reference to the data. Consequently, these new components are likely to be
poor fits to the observed data. This has also been observed by \cite{ge2015distributed}.
In \textit{AV}, large numbers of similar clusters will be proposed from different workers.
However, placement of clusters is random instead of based on similarity.
Therefore, it usually takes multiple iterations for all similar clusters to meet
at a same worker and be merged.
\textit{AV} is fast on NTY dataset because each worker hold only a small portion of components due to sample movement, but many of them cannot be effectively merged. As a result, it results in more than 500 components, while this number is about 160 for our methods and 60 for \textit{SliceMR}.
In \textit{SubC}, new components are generated only in the global split steps.
Hence, more iterations are required to generate enough components to fit the data.
\textit{Hung} performs the same or even a little better than our methods when compared \wrt iterations. However, when compared \wrt runtime, this variant takes 5\% to 10\% longer to converge to the same level of log-likelihood. Because it involves a relatively more expensive procedure.

\begin{figure}[t]
	\includegraphics[width=\columnwidth]{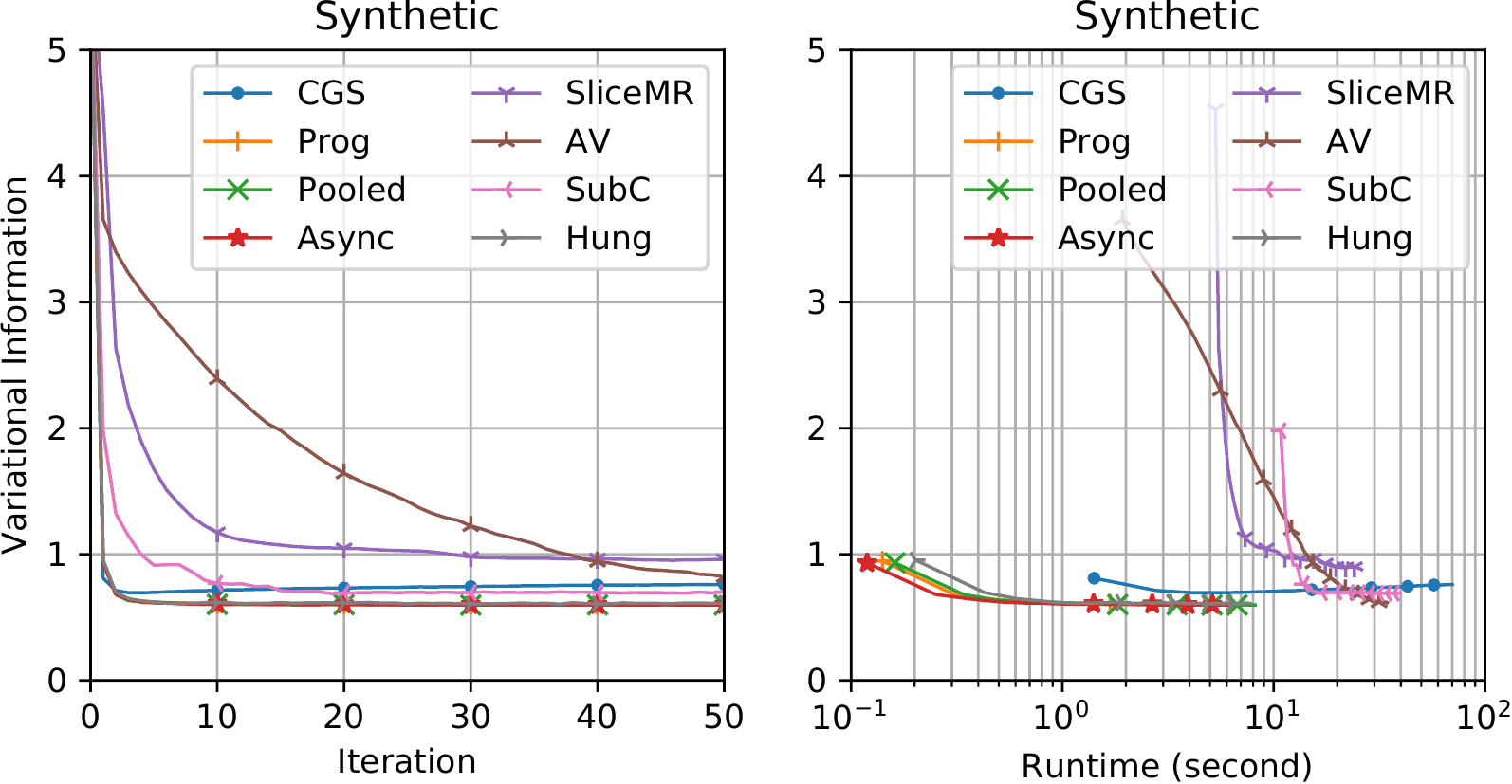}
	\includegraphics[width=\columnwidth]{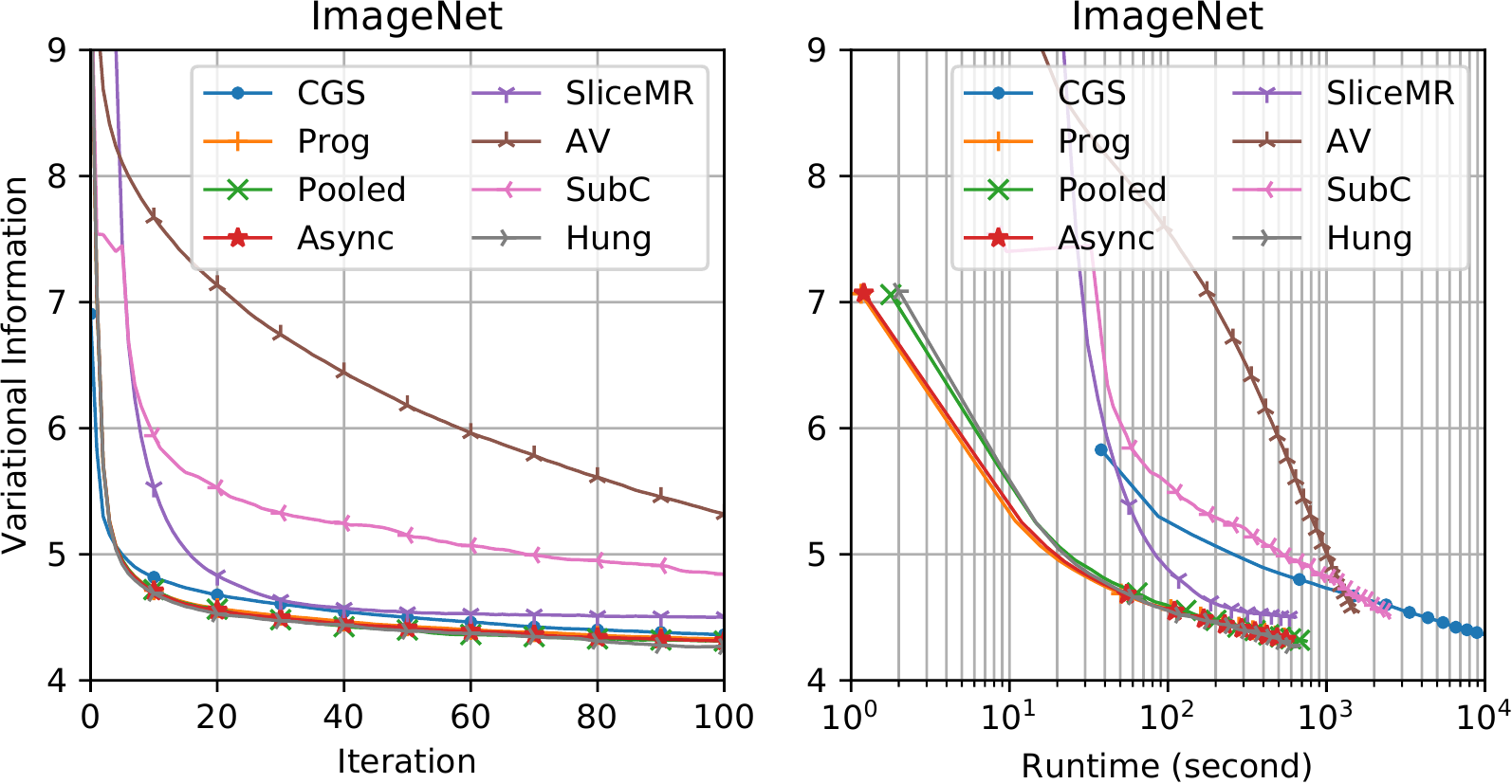}
	\caption{VI \wrt iteration and runtime}
	\label{fig:vi}
\end{figure}

\vspace{-10 pt}
\paragraph{Clustering Performance.}

An important application of mixture modeling is to discover clusters in exploratory data analysis.
Following this practical standpoint, we also tested all algorithms on the first two dataset with provided ground-truths.
Particularly, clustering performance is measured in terms of the
\emph{Variation Information (VI)}~\cite{meilua2003comparing} between
the inferred sample assignments and the ground-truths.
We show the performance metric against both the number of iterations and the clock time in
Figure~\ref{fig:vi}.
Again, compared to CGS,
our methods also achieve the same or even better performance with the same number of iterations,
while taking drastically shorter (about $1/10$ of the CGS runtime on the synthetic dataset,
and about $1/20$ on ImageNet).
\textit{Hung} performs similar as compared to our methods. 
\textit{SliceMR}, \textit{AV} and \textit{SubC} can also achieve reasonable level of VI
on both data sets, but it takes considerably longer.

\vspace{-10 pt}
\paragraph{Communication Cost.}

Communication cost is crucial in distributed computing, especially
when the bandwidth is limited.
We evaluate the communication cost by measuring the number of bytes
communicated and the number of communication times within each iteration,
as shown in Table~\ref{tab:comm_cost}.
Since out methods and \textit{Hung} share the same communication policy, we merge them to one column and fill their the average value in the table.

\begin{table}[t]
	\small
	\begin{subtable}{\columnwidth}
		\centering
		\begin{tabular}{c|cccc}
			& Ours / Hung & SliceMR & AV & SubC \\
			\hline
			\#KBytes & 50.7 & 40.9 & 4819.7 & 2320.4 \\
			\hline
			\#Times & 40 & 40 & 114.6 & 40 \\
		\end{tabular}
		\caption{Synthetic dataset}
	\end{subtable}
	\begin{subtable}{\columnwidth}
		\centering
		\begin{tabular}{c|cccc}
			& Ours / Hung & SliceMR & AV  & SubC \\
			\hline
			\#MBytes & 6.01 & 4.89 & 49.9 & 26.2 \\
			\hline
			\#Times & 40 & 40 & 114 & 40 \\
		\end{tabular}
		\caption{ImageNet dataset}
	\end{subtable}
	\caption{Communication cost}
	\label{tab:comm_cost}
\end{table}

Our methods and \textit{SliceMR} require minimal communication.
In our algorithm, each worker communicates with the master by only two times in each iteration
and the amount of data transfer is proportional to the number of cluster and
the size of sufficient statistics only.
\textit{SliceMR} behaves similarly in this respect -- it only communicates twice per iteration,
transferring only the statistics.
On the contrary, \textit{AV} requires moving clusters among processors in the global MCMC steps.
\textit{SubC} require moving both the labels and the sufficient statistics to the master for
splitting. Thus these two methods require much higher communication costs.

\begin{figure}[t]
	\centering
	\includegraphics[width=\columnwidth]{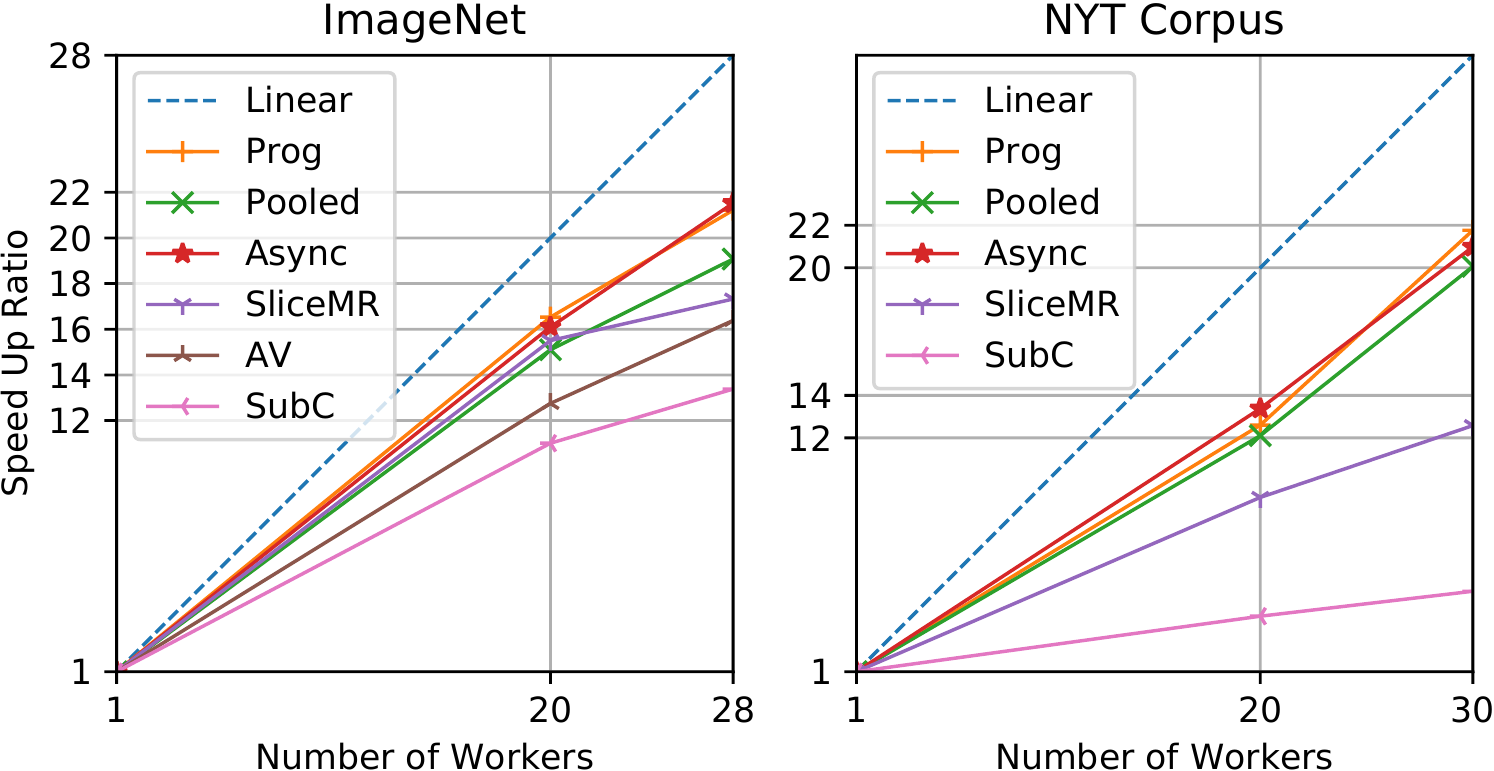}
	\caption{Speed up ratio}
	\label{fig:speed}
\end{figure}

\vspace{-10 pt}
\paragraph{Scalability.}
To study the scalability, we calculated the \emph{speed-up ratio} \wrt~\textit{CGS} for each tested algorithm.
As Figure~\ref{fig:speed} shows, both our \textit{Prog} and \textit{Async} algorithms achieve high scalability,
while the scalability of \textit{Pooled} is little bit lower.
This is expected -- our \textit{Prog} algorithm involves only lightweight global merge steps
while \textit{Pooled} has a heavier MCMC step.
In the \textit{Async} algorithm, consolidation is performed concurrently with local inferences,
thus \textit{Async} can achieve very high scalability as \textit{Prog}.
The scalability of \emph{AV} and \emph{SubC} is considerably poorer, as moving data and label among workers poses significant communication overhead.
Moreover, the unbalanced workload among processors~\cite{gal2014pitfalls} also contributes to the
poor scalability of \emph{AV}.
Note that the high scalability of \textit{SliceMR} is at the expense of the \emph{convergence performance}.
Whereas it achieves nearly $21 \times$ speed up \wrt~its own runtime with a single worker,
the overall performance still leaves a lot to be desired due to slow convergence
(\ie~taking many more iterations than \emph{CGS}).


\section{Conclusions}

We presented a new method for distributed estimation of DPMMs that can work
under both synchronous and asynchronous settings. The method allows workers to
perform local updates and create new components independently through delayed
synchronization, while effectively tackling the issue of component
identification via consolidation. Experimental results on both synthetic and
real-world data clearly show that this method can achieve high scalability
without compromising the convergence rate.


\section*{Acknowledgments}
This work is partially supported by the Big Data Collaboration Research grant from SenseTime Group (CUHK Agreement No. TS1610626) and the General Research Fund (GRF) of Hong Kong (No. 14236516).

\bibliographystyle{named}
\bibliography{Reference}

\end{document}